\documentclass[5p,twocolumn]{elsarticle}

\usepackage{graphicx}
\usepackage{amsmath,amssymb} % define this before the line numbering.
\usepackage{color}

\usepackage[table]{xcolor}
\usepackage{multirow}
\usepackage{soul}
\usepackage{subcaption}
\usepackage{tabularx}
\usepackage[numbers]{natbib} 
\usepackage{epstopdf}
\usepackage{lineno}
\usepackage[hidelinks]{hyperref}
\usepackage{arydshln} % dashline
\usepackage{comment}
\usepackage{enumitem}
\usepackage{array,booktabs,mathtools}

\journal{.}

\biboptions{sort&compress}

\begin{document}  

\begin{frontmatter}
\title{Integrated Neural Network and Machine Vision Approach For Leather Defect Classification}

  \author[add1]{Sze-Teng Liong} %[add1, add3]
  \ead{stliong@fcu.edu.tw}
  \author[add2]{Y.S. Gan}
  \ead{8yeesiang@ntunhs.edu.tw}
    \author[add2]{Yen-Chang Huang}
  \ead{yenchang.huang1@gmail.com}
    \author[add3]{Kun-Hong Liu\corref{cor1}}
  \ead{lkhqz@xmu.edu.cn}
    \author[add4]{Wei-Chuen Yau}
  \ead{wcyau@xmu.edu.my}
  
  \cortext[cor1]{Corresponding author}
  \address[add1]{Department of Electronic Engineering, Feng Chia University, Taichung, Taiwan}
  \address[add2]{Research Center For Healthcare Industry Innovation, National Taipei University of Nursing and Health Sciences, Taipei, Taiwan} 
    \address[add3]{School of Software, Xiamen University, Xiamen, China}
    \address[add4]{School of Electrical and Computing Engineering, Xiamen University Malaysia, Jalan Sunsuria, Bandar Sunsuria, 43900 Sepang, Selangor, Malaysia}

\begin{abstract}
Leather is a type of natural, durable, flexible, soft, supple and pliable material with smooth texture.
It is commonly used as a raw material to manufacture luxury consumer goods for high-end customers.
To ensure good quality control on the leather products, one of the critical processes is the visual inspection step to spot the random defects on the leather surfaces and it is usually conducted by experienced experts.
This paper presents an automatic mechanism to perform the leather defect classification.
In particular, we focus on detecting tick-bite defects on a specific type of calf leather.
Both the handcrafted feature extractors (i.e., edge detectors and statistical approach) and data-driven (i.e., artificial neural network) methods are utilized to represent the leather patches.
Then, multiple classifiers (i.e., decision trees, Support Vector Machines, nearest neighbour and ensemble classifiers) are exploited to determine whether the test sample patches contain defective segments.
Using the proposed method, we managed to get a classification accuracy rate of 84\% from a sample of approximately 2500 pieces of $400\times400$ leather patches.
\end{abstract}


\end{frontmatter}

\section{Introduction}

Leather is a natural product that is made from animal skins (e.g., cow, goat, snake, etc.).
It usually comes with some imperfections and contains blemishes, such as a variety of spots, scratches and irregular color reproduction. %, due to the veterinarian practices and the climates of the living regions.  
%Generally, the price of each piece of leather depends on the size, shape. 
The surface irregularities pose a notable degradation in quality and hence affect its selling price. Therefore, it is vital for manufacturers to detect leather defects during the manufacturing process to ensure the quality of the leather products.

In brief, the steps of processing the hides include:
(1) Soaking: immerse in water and surfactants to remove salt and dirt; 
(2) Unhairing: remove the subcutaneous material and the majority of hair; 
(3) Tanning: convert the raw hides into leather through a chemical treatment; 
(4) Drying: eliminate excess moisture;
(5) Dyeing: custom-color the hides by placing them into the dye drums.
%http://www.townsendleather.com/leather/about-our-leather
Many of the natural defects are not noticeable on the raw hides but appear to be apparent after the tanning process.
Those defective areas with minor damage will be sanded and evened out with fillers to smooth out the surface.

One of the most exhausting procedures during the leather manufacturing process is the defect determination process. 
A typical exercise in quality control during the production is to perform rigorous manual inspection on the same piece of leather several times, using different viewing angles and distances.
%Nevertheless, a single piece of leather can be as large as 3m $\times$ 3m.
To date, the most common solution is to manually mark the defect areas, and then rank the defects digitally, based on the severity level (i.e., minor, major, critical, etc.) and the defect types (i.e., cuts, tick bites, wrinkle, scabies, etc.).
However, the process of the human inspection is expensive, time consuming, and subjective.
In addition, it is always prone to human error as it requires a high level of concentration and might lead to labour fatigue.
Therefore, there is a necessity to develop an automatic vision-based solution in order to reduce manual intervention in this specific process.

There are several image processing solutions proposed in the literature.
For instance, Georgieva et al.~\cite{georgieva2003identification} suggested to determine the leather defects by computing a chi-squared ($\chi^2$) distance between the gray-level histograms from a reference image and the test images.
The identification of the leather defect is based on the similarity or dissimilarity of the color histograms. 
However, there is lack of qualitative and quantitative experimental demonstration in the paper. 
Therefore, the effectiveness of the proposed method seems to be unknown.

On the other hand, Branca et al.~\cite{branca1996automated} proposed a leather defect detection method using a Gaussian filter and its oriented texture.
A Gaussian convolution is first applied on the leather image which serves as the smoothing operator to reduce the image noise.
Then, the orientation flow field details are computed to derive the local gradient and its corresponding length for each neighborhood pixel.
A neural network is trained by treating the orientation vector field as the input and representing them as sets of projection coefficients.
Finally, the defects will be classified depending on these coefficients.
The paper illustrates the results of nine samples which highlight the segmented areas as the defective regions. 
However, there are no performance metrics involved to quantify and verify the correctness of the proposed algorithm.

Amorim et al.~\cite{amorim2010attributes} demonstrated several attributes reduction methods to perform leather defect classification.
They utilize several FisherFace feature reduction techniques to represent the details of the leather images by projecting the attributes vectors onto a subspace.
The initial length of the features is 4202 per sample, which include the attributes of color details (hue, saturation, brightness, red, green and blue), histograms of the color, co-occurrence matrix, Gabor filters and the original pixels.
After applying the discriminant analysis techniques, the feature length is reduced to 160 per sample.
They tested on 2000 samples that contain eight classes, namely background, no-defect, hot-iron marks, ticks, open cuts, closed cuts, scabies and botfly larvae.
The classifier types include C4.5, $k$-Nearest Neighbors (kNN), Naive Bayes and Support Vector Machine (SVM). 
The highest classification accuracy obtained is $\sim$88\% for wet blue images and $\sim$92\% for raw hide images. 

A defect classification process on goat leather samples was carried out in~\cite{pereiraclassification}.
A combination of features from Gray level Co-Occurrence Matrix (GLCM), Local Binary Pattern (LBP) and Pixel Intensity Analyzer (PIA) are formed.
Then, classifiers such as kNN, MLM (Minimal Learning Machine), ELM (Extreme Learning Machine), and SVM are adopted independently and tested on the features extracted.
The dataset collected comprises of 1874 samples with 11 classes (normal, wire risk, poor conservation, sign, bladder, scabies, mosquito bite, scar, rufa, vegetable fat and hole).
The ground-truth defects are annotated by two leather classifier specialists who have undergone one month's training.
As a result, an accuracy of 89\% is exhibited when using the LBP feature descriptor on the SVM classifier.

There are very few works in the literature exploiting deep learning techniques in analyzing the leather types.
One of the recent works that applied a pre-trained Convolutional Neural Network (CNN) was conducted by Winiarti et al.~\cite{winiarti2018pre}.
Using approximately 3000 leather sample images from the five types of leather (i.e., monitor lizard, crocodile, sheep, goat, and cow skin), they achieved a 99.9\% classification accuracy.
Succinctly, they performed transfer learning process using AlexNet on 1000 leather images, with 200 leather images for each category.
The paper also illustrates that there might be differences in the texture within the same category.
Thus, it shows that deep neural networks perform well in leather classification tasks.

In a recent work on automated defect segmentation, Liong et al.~\cite{liong2019automatic} introduced a series of algorithms to predict tick-bite defects on leather samples .
Different from the conventional methods that capture the leather sample manually, ~\cite{liong2019automatic} elicits all the image data using a robot arm and draws the defect region with a chalk using the same robot arm.
One of the state-of-the-art models for instance segmentation, namely, Mask Region-based Convolutional Neural Network (Mask R-CNN) is employed and fine tuned with 84 defective images to learn the local features of the leather samples.
The proposed method exhibits an accuracy of 70\% on 500 testing images.
It should be noted that this segmentation task differs from the classification task, where the former localizes the defect region and the latter distinguishes the type of the defect on a leather sample.

The objective of this paper is to differentiate defective and non-defective leather samples, particularly focusing on the tick bite defect type.
Both the handcrafted and data-driven feature descriptors are employed to extract the local information of the leather patches.
Specifically, the handcrafted features include edge detector, 2D convolution and the histogram of the pixel values.
Next, the feature sets are fed into several dominant machine classifier models independently, to predict the leather's defective status.
The classifiers are decision tree, SVM, $k$-NN and a set of ensemble classifiers.
As for the neural network framework, it is designed to discover the intricate structure in the leather patches.
A simple Artificial Neural Network (ANN) is employed to categorize the testing data by providing it with the training input data and output label  prior to building and training the machine learning model.

In summary, the contributions of this research work are listed as follows:
\begin{enumerate}
    \item Proposal of ANN-learned and handcrafted features extractors independently on each leather sample patch.
    \item Demonstration of the scalability of the proposed algorithm by evaluating them on a variety of machine learning classifiers.
    \item Thorough experimental assessment and analysis are conducted on approximately 2500 images.
    \item Both the qualitative and quantitative results are reported and show that the proposed approach achieves promising classification performance.
    
\end{enumerate}

The rest of the paper is arranged as follows. 
Section~\ref{sec:proposed} explains the algorithms
of the proposed defect classification system for leather in detail, specifically, the theoretical definitions of the image processing techniques utilized.
The experiment configuration such as the database, performance metrics, and the parameters values in the algorithm are presented in Section~\ref{sec:experiment}.
The results are discussed in Section~\ref{sec:result} and finally, the conclusions are then drawn in Section~\ref{sec:conclusion}, together with some suggestions for further research studies.

\section{Proposed Method}
\label{sec:proposed}

\begin{figure*} []
	\centering
	\includegraphics[width = 0.8\linewidth]{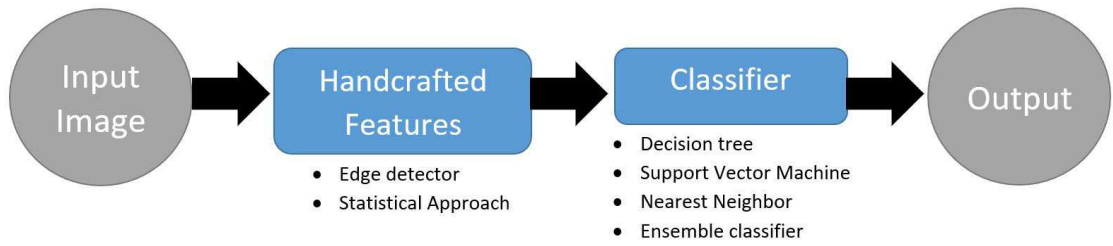}
	\caption{Overview of the proposed leather defect classification system, which consists of the feature extraction and classification stages. }
	\label{fig:flow}
\end{figure*}

There are two types of handcrafted features extraction methods utilized in this study, namely, the edge detector and the statistical approach.
Then, the features extracted from the training data are modeled and generalized by several classifiers to later predict the output class of the testing data. 
The flowchart of the handcrafted feature extraction and classification is illustrated in Figure~\ref{fig:flow}.
In our proposed method, we investigate some well-known edge detection operators to estimate the boundaries of objects in an image by using the discontinuities in colour intensity.
They include Prewitt~\cite{prewitt1970object}, Roberts~\cite{roberts1963machine}, Sobel~\cite{lyvers1988precision}, Laplacian of Gaussian (LoG)~\cite{marr1980theory}, Canny~\cite{canny1987computational} and Approximation Canny (ApproxCanny)~\cite{matlab} operators. 
On the other hand, for the statistical approach, we focus on the histogram of pixel intensity values (HPIV), Histogram of Oriented Gradient (HOG)~\cite{dalal2005histograms} and Local Binary Pattern (LBP)~\cite{ojala2002multiresolution}.
As for the classification stage, state-of-the-art supervised classifiers are employed, such as decision tree~\cite{safavian1991survey}, discriminant analysis~\cite{klecka1980discriminant}, SVM~\cite{suykens1999least}, Nearest Neighbor (NN)~\cite{cover1967nearest} and some ensemble classifiers.
The details of mathematical derivations of aforementioned handcrafted feature extractors and classifiers are elaborated in Section~\ref{subsec:handcrafted} and Section~\ref{subsec:classifier}, respectively.

\begin{figure*} []
	\centering
	\includegraphics[width = 0.6\linewidth]{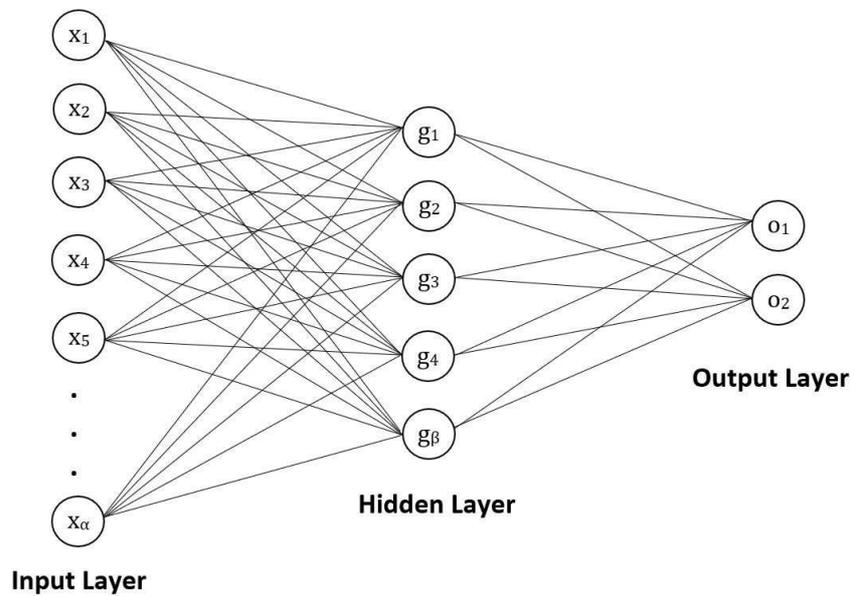}
	\caption{Illustration of an Artificial Neural Network (ANN)}
	\label{fig:flow2}
\end{figure*}

Besides that, we attempt to design and train the ANN architecture, which is also considered as a data-driven predictive model. 
Note that a single ANN architecture can serve as both feature encoder and classifier.
The visualization of the ANN is shown in Figure~\ref{fig:flow2} and its details are described in Section~\ref{subsec:ANN}.

%%%%%%%%%%%%%%%%%%%%%%%%%%%%%%%%%%%%%
\subsection{Handcrafted Features}
\label{subsec:handcrafted}

\subsubsection{Edge Detector}
In image processing, variants of mathematical methods aim to identify the changes of brightness in an image. 
The discontinuity points at which the image brightness changes the most are then modelled as edges of the objects.
This is the fundamental concept of the edge detection operation.
Ideally, an edge detector will output a set of connected curves that indicate the boundaries of objects in an image.
Meanwhile, it significantly reduces the dimensionality of feature vectors by eliminating irrelevant information, or noise, in an image.
In the discrete case, the directional derivatives of both the horizontal and vertical directions are approximated by simple finite differences~\cite{marr1980theory}.
There are several edge detectors based on the first and second order rates of change, as discussed in the following:

\begin{enumerate}[label=(\alph*)]
	\item  Prewitt operator\\
	It computes an approximation of the gradient of the image intensity function. 
	At each point in the image, the Prewitt operator uses a small integer-valued kernel either in horizontal or vertical directions to convolve the image, resulting in the corresponding gradient vector. % or the norm of this vector. 
	%Mathematically, the horizontal and vertical derivative approximations based on Prewitt operator  
	Commonly, the 3$\times$3 kernel for this horizontal and vertical derivative approximation are denoted as $G_x$ and $G_y$, respectively:

	\begin{equation} 
			G_x =  \begin{bmatrix}
				1  & 0 & -1 \\
				1 & 0 & -1 \\
				1 & 0 & -1
			\end{bmatrix}, 
			G_y =  \begin{bmatrix}
				1  & 1 & 1 \\
				0 & 0 & 0 \\
				-1 & -1 & -1
			\end{bmatrix} 
	\end{equation}
	
	At each sampling point in the image, the gradient approximations can be merged, and molded to the gradient's magnitude ($\rho$) and gradient's direction ($\theta$):
	
	\begin{equation} \label{mag}
		\rho = \sqrt{G_x^2 + G_y^2},
	\end{equation}
	
	\begin{equation} \label{orien}
		\theta = tan^{-1} \left(\frac{G_y}{G_x}\right).
	\end{equation}
	
	For example, the zero value of $\theta$ of a vertical edge, that is darker on the right side.
	
	\item  Roberts operator \\
	The Roberts operator performs a quick spatial gradient approximation on an image by highlighting the regions of high spatial gradient through a discrete differentiation process.
	It is achieved by computing the sum of squares of the differences between diagonally adjacent pixels. 
	The idea of Roberts operator is to apply a pair of kernel masks on the image, whereby the second mask is simply the first rotated by 90$^\circ$.  
	Both the horizontal and vertical derivative approximations based on this Roberts operator by using a 3$\times$3 kernel, are defined below:
	
	\begin{equation}
			G_x =  \begin{bmatrix}
				0  & 0 & 0 \\
				-1 & 1 & 0 \\
				0 & 0 & 0
			\end{bmatrix}, 
			G_y =  \begin{bmatrix}
				0  & 0 & 0 \\
				0 & 1 & 0 \\
				0 & -1 & 0
			\end{bmatrix}.
	\end{equation}
	
	The resultant gradient's magnitude ($\rho$) and gradient's direction ($\theta$) are similar as Equation~\ref{mag} and Equation~\ref{orien}, respectively.
	
	\item Sobel operator \\
	Similar to Roberts operator, the mask for Sobel operator is designed to respond maximally to edges running vertically and horizontally, relative to the pixel grid, that is one mask for each of the two perpendicular orientations.   
	The horizontal and vertical derivative approximations by using the 3$\times$3 kernel are:
	
	\begin{equation}
			G_x =  \begin{bmatrix}
				-1  & 0 & 1 \\
				-2 & 0 & 2 \\
				-1 & 0 & 1
			\end{bmatrix},
			G_y =  \begin{bmatrix}
				1  & 2 & 1 \\
				0 & 0 & 0 \\
				-1 & -2 & -1
			\end{bmatrix}.  
	\end{equation}
	
	The resultant gradient's magnitude ($\rho$) is similar as Equation~\ref{mag}.
	However, its gradient's direction ($\theta$) is expressed as: 
	
	\begin{equation}
		\theta = tan^{-1}\left(\frac{G_y}{G_x} \right) - \frac{3\pi}{4}.
	\end{equation}
	
	\item Laplacian of Gaussian (LoG)\\
	LoG is a blob detection method that is intended to detect regions in a digital image that has different properties, such as brightness or color by comparing with its neighbor pixels.
	Since the derivative filters (i.e., Prewitt, Roberts and Sobel operators) are very sensitive to noise, LoG is introduced to overcome this issue by smoothing the image (i.e., using a Gaussian filter) prior to applying the Laplacian operation.  
	The equation defined for LoG is given as follows:
	
	\begin{equation}
		LoG(x,y) = -\frac{1}{\pi\sigma^4} \left[1 - \frac{x^2+y^2}{2 \sigma^2} \right] \exp{-\frac{x^2+y^2}{2\sigma^2}}.
	\end{equation}
	
	\item Canny operator\\
	This is similar to LoG, in that it overcomes the noise susceptibility problem.
	In brief, the process of Canny edge detection algorithm has been devised using these five steps:
	
	\begin{enumerate}[label=(\roman*)]
		\item A Gaussian filter is applied to eliminate the image noise and reduce the fine details.
		\item The intensity gradients of the image is derived.  
		For instance, the edge detector operators such as Prewitt, Robert, Sobel can be adopted to acquire the gradients. 
		\item A non-maximum suppression is exploited to get rid of spurious response to edge detection.
		\item The values of the thresholds in the double-threshold method are defined to determine potential edges.
		\item The edges are tracked by hysteresis, which is to minimize all the \textit{weak} edges that are not connected to \textit{strong} edges.
	\end{enumerate}
	
	\item Approximation Canny operator (ApproxCanny)\\
	A simple practical approximation of the Canny operator is developed by applying Gaussian filter with different-sized filters to smooth the image.  
	The outputted smooth image is then put through a gradient operator ($G_x$, $G_y$) to obtain the gradient's magnitude and orientation.
	ApproxCanny is a relatively sparse estimation compared to Canny operator, which allows for less computation time but less precise edge detection on images with tiny details.
\end{enumerate}

%\subsubsection{2D Convolution}

%\begin{bmatrix}
%    -1  & 0 & 1 \\
%    -2 & 0 & 2 \\
%    -1 & 0 & 1
%\end{bmatrix}

%\paragraph{conv2 [ -1 0 1; -2 0 2; -1 0 1]}

%\paragraph{conv2 [1 2 1; 0 0 0; -1 -2 -1]}

%\paragraph{conv2 [ -1 0 1; -2 0 2; -1 0 1] + [1 2 1; 0 0 0; -1 -2 -1]}

\subsubsection{Statistical Approach}

\begin{enumerate} [label=(\alph*)]
	\item  Histogram of Pixel Intensity Values (HPIV)\\
	Histogram is a classic tool to graphically represent the pixel intensities of an image in a compact way.
	HPIV indicates the distribution of the number of pixels according to their intensity value.
	For instance, in an 8-bit grayscale image, there are total 256 (i.e., 2$^8$) different intensity values which range from 0 to 255.
	The formal definition for the histogram is denoted as:
	
	\begin{equation}
		h(i) = Card\{(u,v) | I(u,v) = i\},
	\end{equation}
	where $Card$ is the Cardinality of a set and $I(u,v)$ refers to the intensity at a sampling coordinate $(u,v)$ and $i$ is the intensity value.
	
	\item Histogram of Oriented Gradient (HOG)\\
	HOG is a frequency histogram gradient of orientation for each local patch of an image. 
	On a dense grid of evenly spaced cells, HOG computes the frequency of the intensity gradients or edge directions to encode the local appearance features and the shape of an object.
	Next, the histogram of gradient directions within the connected cells are concatenated to form the resultant rich feature vector.  
	The advantages of the HOG descriptor are: 
	(1) Easy and quick to compute; 
	(2) Ability to encode local shape information, and;
	(3) Invariant to geometric and photometric transformations.
	The following lists the steps of the HOG algorithm:
	
	\begin{enumerate}[label=(\roman*)]
		\item Gradient Image Creation \\
		The movements of energy from left to right and up to down are computed.
		This can be achieved by applying the image filtering method that contains both the horizontal and vertical kernels.  
		For example, the $[-1~0~1]$ and $[-1~0~1]^T$ kernels.  
		Then, Equation~\ref{mag} and Equation~\ref{orien} are adopted to extract the magnitude and orientation of each pixel.  
		Here, the constant colored background is removed from the image, but important outlines are kept. 
		For the color images, the maximum magnitude of gradient from the three channels are captured together with their corresponding angles.
		
		\item HOG Computation in $m \times n$ Cells \\
		In order to describe an image in a compact representation with less noise, an image can be divided into $m \times n$ cells with a 9-bin histogram.
		The 9 bins corresponds to the angles $0, 20, 40, 60, ..., 180$.
		
		\item Cell's Blocks Normalization\\
		The magnitude information may be susceptible to the changes in illumination and contrast. In order to resolve this issue, a normalization is applied locally for each block. 
		Finally, the feature vector is enriched by concatenating the histograms for each normalized block.
	\end{enumerate}

	\item Local Binary Pattern (LBP)\\
	LBP is a kind of feature extractor that generates the local representation feature vector for object detection.
	It serves to form a feature vector by comparing each pixel with its surrounding neighbourhood of pixels.
	The procedure to derive LBP features is as follows:
	
	\begin{enumerate} [label=(\roman*)]
		\item The region of interest is divided into $m \times n$ cells.
		\item For each pixel in a cell, the intensity value of the center pixel (x$_c$ , y$_c$) is compared to its circular surrounding pixels using a thresholding technique:
		\begin{equation}
			\label{eq:lbp}
			LBS_{P,R} = \sum\limits_{p=0}^{P-1} s(g_c - g_p) , s(x) = \begin{dcases*}
				1,  & x $>$ 0\\
				0, & x $\leq$ 0
			\end{dcases*}
		\end{equation}
		where $P$ refers to the number of neighbouring points surrounding the center pixel. $(P,R)$ is the neighbourhood of $P$ sampling points evenly spaced on a circle of radius $R$. $g_c$ is the gray intensity value of the center pixel and $g_p$ are the $P$ gray intensity values of the points in the neighbourhood.
		\item This produces a set of $P$-digit binary numbers, which is usually then converted to a decimal number.
		\item A histogram is formed from the LBP image and serves as the final feature vector.
	\end{enumerate}     
	
\end{enumerate}
%%%%%%%%%%%%%%%%%%%%%%%%%%%%%%%%%%%%%
\subsection{Classifier}
\label{subsec:classifier}
After obtaining the feature vectors from the feature descriptions described in  the previous section, they are then fed into the classifier to perform the object class recognition task. 
We select some widely known classifiers, notably, decision tree, SVM, NN and ensemble classifier.
In general, the functions for each of them are:
\begin{enumerate}
	\item Decision Tree: Interpret the class of the input data by outlining all the possible consequences.
	A decision tree is comprised of root, nodes, branches and leaves.
	The response is made by following the decision from the root node to the leaf node.
	\item SVM: Utilize a kernel transform on the feature vector to obtain an optimal boundary between the possible outcomes.  
	\item NN: Determine the class of the input data by selecting the number of majority votes from its neighbours. 
	The most common NN technique is $k$-NN, whereby when $k=1$, the predicted output will be categorized into the class according to a single neighbor.
	\item Ensemble classifier: Combines several weak classifiers into a ensemble model, in order to integrate their discrimination capabilities.
\end{enumerate}
The details for each type of classifier are summarized in Table~\ref{table:classifier}.

\begin{table*}[t!]
	\begin{center}
		\caption{Description for each supervised classifier}
		\label{table:classifier}
		\begin{tabularx}{\linewidth}{lcX}
			
			\hline
			\noalign{\smallskip}
			\multicolumn{2}{c}{Classifier} 
			& Description\\

			\hline
			\multirow{3}{*}{Tree}
			& Coarse Tree 
			& Each leaf node splits into maximum 4 child nodes to allow coarse distinctions between classes. 
			\\
			
			& Medium Tree
			& Each leaf node splits into maximum 20 child nodes to allow finer distinctions between classes. 
			\\
			
			& Fine Tree
			& Each leaf node splits into maximum 100 child nodes to allow many fine distinctions between classes. 
			\\
			
			\hline
			\noalign{\smallskip}
			
			\multirow{6}{*}{SVM}
			& Linear
			& Creates a linear separation between classes and has low model flexibility.
			\\
			
			& Quadratic
			& Creates a quadratic function to separate the data between classes and has medium model flexibility.
			\\
			
			& Cubic
			& Creates a cubic polynomial function to separate the data between classes and has medium model flexibility.
			\\

			& Fine Gaussian 
			& The kernel scale is fixed to ${\frac{\sqrt P}{4}}$ to create high distinctions between classes, where $P$ is the number of predictors.
			\\
			
			& Medium Gaussian 
			& The kernel scale is fixed to ${\sqrt P}$ to create medium distinctions between classes.
			\\
			
			& Coarse Gaussian 
			& The kernel scale is fixed to 4$\sqrt P$ to create coarse  distinctions between classes.
			\\
			
			\hline
			\noalign{\smallskip}
			
			\multirow{6}{*}{$k$NN}
			& Fine 
			& The number of neighbors is set to 1 and employs a euclidean metric.
			\\
			
			& Medium 
			& The number of neighbors is set to 10 and employs a euclidean metric.
			\\
			
			& Coarse
			& The number of neighbors is set to 100 and employs a euclidean metric.
			\\

			& Cosine 
			& The number of neighbors is set to 10 and employs a Cosine distance metric.
			\\
			
			& Cubic 
			& The number of neighbors is set to 10 and employs a cubic distance metric.
			\\
			
			& Weighted 
			& The number of neighbors is set to 10 and employs distance-based weighting.
			\\
			
			\hline
			\noalign{\smallskip}
			
			\multirow{5}{*}{Ensemble}
			& Boosted Tree 
			& Combines AdaBoost and decision tree learners.
			\\
			
			& Bagged Tree  
			& Combines Random Forest and decision tree learners.
			\\
			
			& Subspace Discriminant 
			& Combines Subspace and discriminant learners.
			\\

			& Subspace $k$NN 
			& Combines Subspace and NN learners.
			\\
			
			& RUSBoosted Tree
			& Combines RUSBoost and decision tree learners.
			\\
			
			\hline
			\noalign{\smallskip}
			
		\end{tabularx}
	\end{center}
\end{table*}

\subsection{Artificial Neural Network (ANN) Learning Features}
\label{subsec:ANN}
ANN has demonstrated its feasibility in the pattern classification task.
Owing to its ability to correlate complex relationships into a model and its remarkable generalization capability, it can be deployed in several applications such as handwritten text recognition~\cite{espana2011improving}, weather forecasting~\cite{taylor2002neural}, financial economics~\cite{li2010applications} and even agricultural land assessment~\cite{wang1994use}.

Basically, ANN is devised into three layers, viz., the input, hidden and output layers, as shown in Figure~\ref{fig:flow2}.
The number of neurons of the input layer depends on the feature vectors extracted from the input data, whereas the number of neurons for hidden layer is subjective as it relies on the objective of the problem and the complexity of the function. 
In general, the neurons in both the hidden and output layers are the sigmoid activations.  
The output of the ANN can be described by following equation:

\begin{equation}
	y_{output} = \frac{1}{1 + \exp^{-(b_2 + W_2 (max(0, b_1 + W_1 \ast X_n)))}} 
\end{equation}

\noindent where $W_1, W_2, b_1, b_2$ are weights and biases in the network and $X_n$ is the input value.
In order to optimize the performance of the ANN, the training process is usually carried out based on the Adam optimization algorithm~\cite{kingma2014adam}, to adaptively adjust the weights and biases in the network.  

\section{Performance Metrics and Experiment Setup}
\label{sec:experiment}
\subsection{Database}
\begin{figure} [t!]
    \centering
    \includegraphics[width = 1\linewidth]{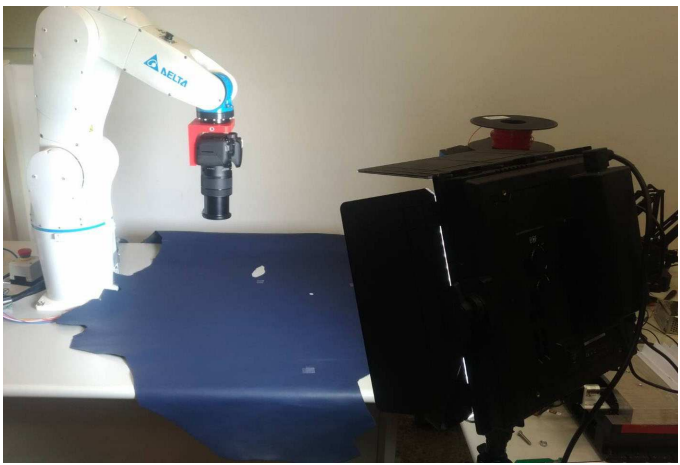}
    \caption{Illustration of the experimental setup}
    \label{fig:setup}
\end{figure}

The dataset contains 2378 images of the sample patches on a piece of leather that is approximately 90$\times$60 $mm^2$ in width$\times$length.
Among them, 475 images have at least one tick bite defect, while 1903 images do not contain any defects. 
All the images are acquired with a robot arm; a six-axis articulated robot DRV70L from Delta, which is able to embed 5kg of payload.
The robot arm is equipped with a Canon 77D camera fitted with a 135mm focal length lens to meet our resolution requirement of 37.5$\mu$m/pixel.
The usage of the robot arm is to ensure constant, consistent and accurate distance from the leather object to the camera.
Thus, the robot arm will move vertically and horizontally to capture leather images with spatial resolution of 2400$\times$1600 pixels$^2$.
DOF D1296 Ultra High Power LED light is utilized, whereby it adopts 1296 high-quality LED light beads of extra-large luminous chip and the illumination is up to 12400 lux.
This is to ensure a uniform light distribution and flicker-free light environment.
An illustration of the experimental setup is shown in Figure~\ref{fig:setup}, which contain the robotic arm, camera, LED light source and the leather.

\begin{figure*} [t!]
    \centering
    \includegraphics[width = 1\linewidth]{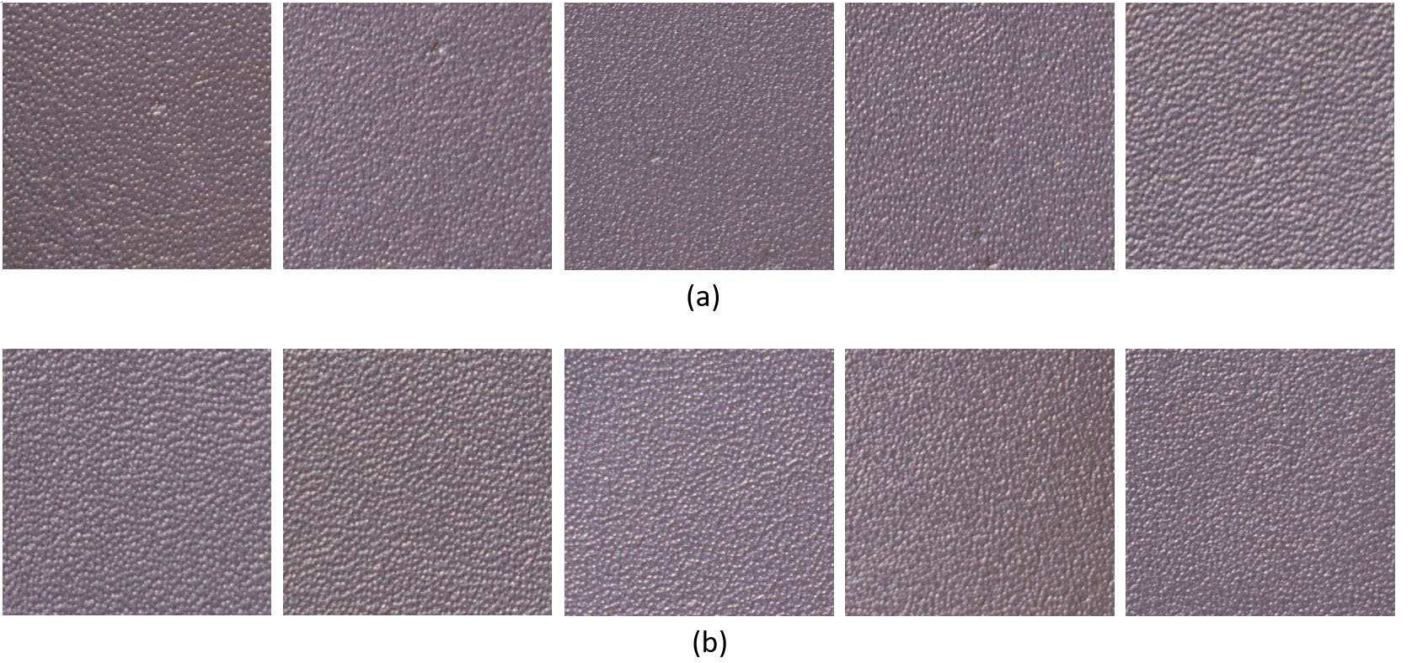}
    \caption{Example for the leather patches (a) with defect, and; (b) without defect}
    \label{fig:sample}
\end{figure*}

\begin{figure} [t!]
    \centering
    \includegraphics[width = 0.7\linewidth]{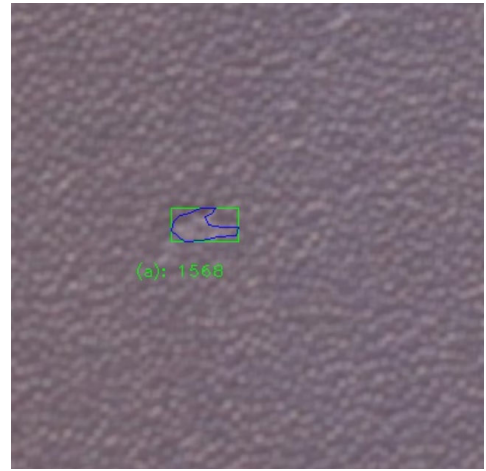}
    \caption{Estimation of the surface area of the defect using a bounding box}
    \label{fig:sampleArea}
\end{figure}

\begin{figure} [t!]
    \centering
    \includegraphics[width = 1\linewidth]{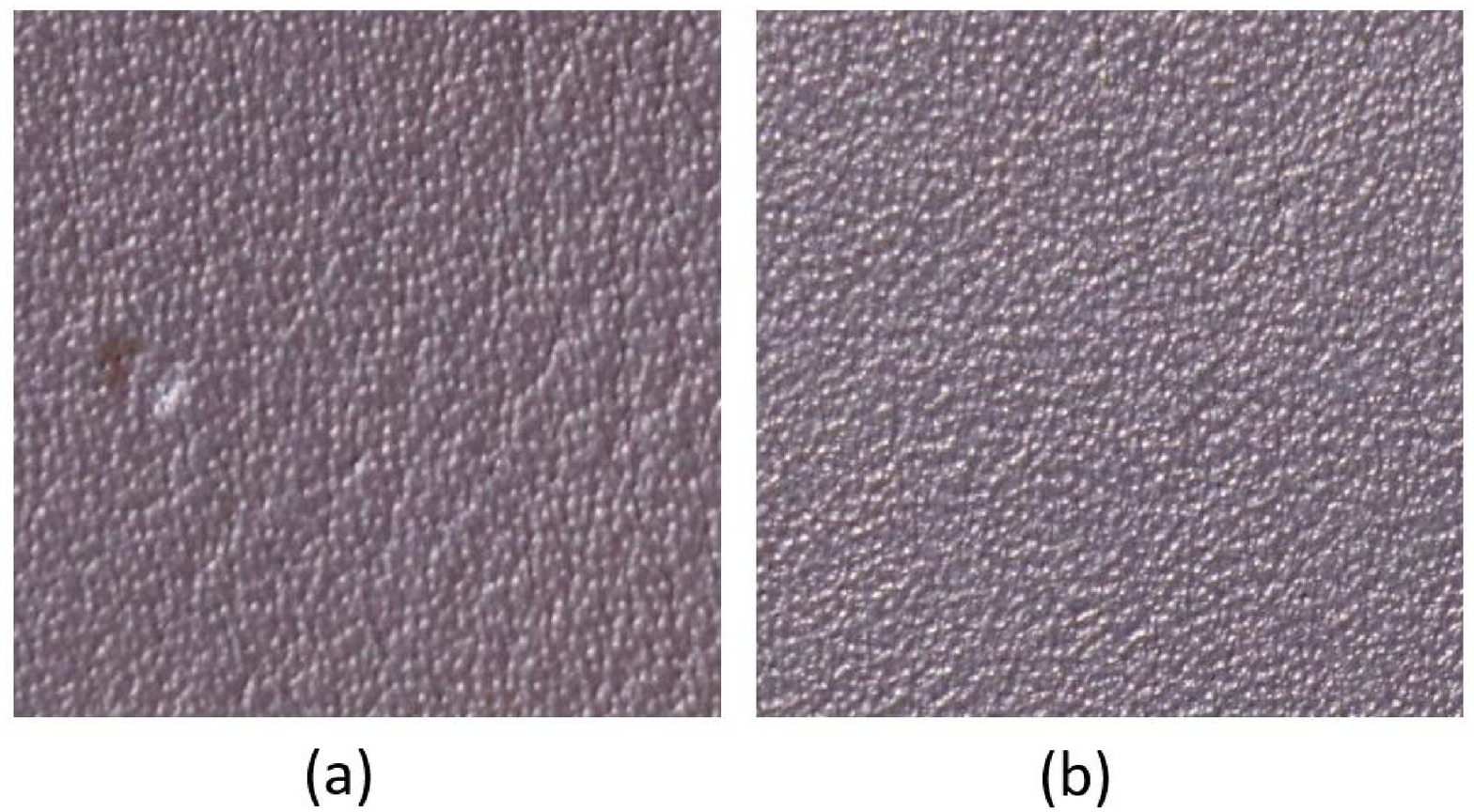}
    \caption{The (a) largest and (b) smallest defect size in the dataset.}
    \label{fig:small_large}
\end{figure}

Figure~\ref{fig:sample} shows the samples for the leather patches with defects and without defects.
It should be noted that some of the images contain more than one defect.
The statistical analysis for the tick bite defects is presented in Table~\ref{table:area}, which includes the width, height and surface area of the defect. 
The area for the smallest defect is $\sim$0.042mm$^2$, whereas the largest defect is $\sim$4.493mm$^2$.
Figure~\ref{fig:sampleArea} illustrates a bounding box around the defect with an estimate of its size.
The largest and the smallest defect size in the dataset are shown in Figure~\ref{fig:small_large}

\setlength{\tabcolsep}{5pt}
 \begin{table}[tb]
 \begin{center}
 \caption{Statistical analysis for the tick bite defects on the leather patches}
 \label{table:area}
 \begin{tabular}{lcccc}
 \noalign{\smallskip}
 \cline{2-5}
\noalign{\smallskip}

 & x-axis 
 & y-axis 
 & Area 
 & Area\\\

 & (pixel)
 & (pixel)
 & (pixel$^2$)
 & (mm$^2$)\\

 \hline
\noalign{\smallskip}
Minimum 
 & 6
 & 5
 & 30 
 & 0.0422\\
 
\noalign{\smallskip}
First quartile
 & 16
 & 16
 & 272 
 & 0.3825\\
 
\noalign{\smallskip}
Median
 & 20
 & 20
 & 396 
 & 0.5569\\

\noalign{\smallskip}
Third quartile
 & 26
 & 25
 & 575 
 & 0.8086\\

\noalign{\smallskip}
Maximum
 & 65
 & 71
 & 3195
 & 4.4930 \\
 
\noalign{\smallskip}
Mean
 & 21.56
 & 20.86
 & 480.44 
 & 0.6756\\
 
\noalign{\smallskip}
Standard deviation
 & 8.22
 & 7.58
 & 347.29 
 & 0.4884\\
 \hline

 \end{tabular}
 \end{center}
 \end{table}

\subsection{Experiment Configuration}
The experiments were carried out using MATLAB 2018b on an Intel Core i7-8700K 3.70 GHz processor, RAM 48.0 GB, GPU NVIDIA GeForce GTX 1080 Ti.
All the leather images are first resized to 40$\times$40 before encoding the features.
This is to reduce the computational complexity and enhance the execution speed.
For the handcrafted feature extractors (i.e., edge detector and statistical approach) and the classifiers (i.e., decision tree, discriminant analysis, SVM, NN, ensembler classifier), a 5-fold cross validation is applied.
This is to evaluate the performance of the machine learning models on unseen data. 
For the HOG feature descriptor, we evaluated the cell size of the default values [8 8] and [10 10].
As for the LBP, the cell sizes of [8 8], [16 16] and [32 32] are tested.

On the other hand, for the shallow neural network (i.e., ANN),
we use a three-layer network with $x$ input and $o$ output neurons.
An investigation is conducted by varying the number of neurons in the hidden layers, $g$.
A tan-sigmoid transfer function is utilized in the hidden layer, and a softmax transfer function in the output layer
In addition, a few types of train/ test partitions on the 2376 samples have been tested. 
For instance, the random division of train/ test splits are 70/30, 75/25, 80/20, 85/ 15, 90/10 and 95/5.

\subsection{Performance Metrics}

There are 2 classification classes (defect or no defect) in the prediction. 
Thus the four outcomes can be formulated in a $2\times2$ confusion matrix, that is comprised of true positive (TP), true negative (TN), false negative (FN) and false positive (FP).
In brief, TP indicates that the model correctly predicts that the image contains defects.
TN means that the model correctly predicts that there is no defect.
FN is the model fails to detect the defect, while in fact there is a defect.
FP indicates that the model detects a defect, but there is none in the image.
The evaluation metric to validate the effectiveness of the proposed approach is accuracy:

\begin{equation}
\text{Accuracy} := \frac{\text{TP + TN}}{\text{TP + FP + TN + FN}}
\end{equation}

\subsection{Visualization after Applying Edge Detection Filters}
\begin{figure} [t!]
    \centering
    \includegraphics[width = 1\linewidth]{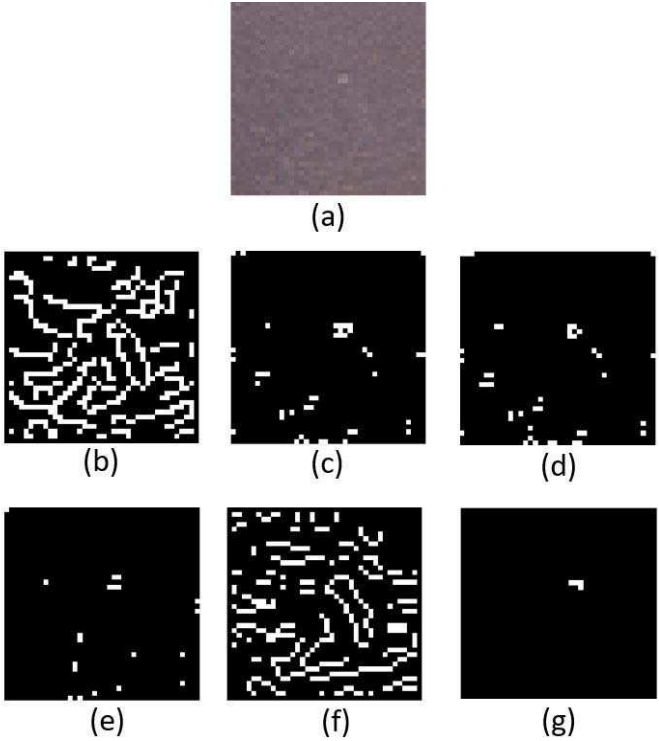}
    \caption{Edge detection effect on a defective sample: (a) Original image ;(b) Canny (c) Prewitt; (d) Sobel; (e) Roberts ;(f) LoG, and (g) ApproxCanny}
    \label{fig:edge}
\end{figure}

The qualitative comparison after performing the edge detection step is shown in Figure~\ref{fig:edge}.
For all these six edge detector methods,  both the horizontal and vertical directions of the edges are detected.
It is observed in Figure~\ref{fig:edge}, there is significant difference between the output images.
Among them, only \textit{ApproxCanny} operator is able to precisely highlight the defect in this specific sample.

\section{Result and Discussion }
\label{sec:result}
Table~\ref{table:resultEdge} and Table~\ref{table:resultHist} show the classification results by employing various types of edge detectors and statistical approaches as feature extractors.
The highest classification accuracy achieved with edge detector is 83.50\%, which is yielded by \textit{ApproxCanny}.
Table~\ref{table:resultEdge} shows promising results, with an average accuracy of more than 75\% for all the classifiers.
\textit{ApproxCanny} edge detector is an approximate version of the \textit{Canny} edge detector, where the \textit{Canny} is relatively precise at edge positioning.
However, since \textit{Canny} is quite sensitive to noise, it may falsely detect many fine edges, as shown in Figure~\ref{fig:edge}(b).
From the original image shown in Figure~\ref{fig:edge}(a), the leather has fine grain structure, hence, \textit{Canny} edge detector probably is not an optimal detector in this experiment.
In contrast, \textit{ApproxCanny} has high execution speed but less precise detection, which best suits the requirement in our case, as it can be seen that Figure~\ref{fig:edge}(g) obviously depicts the detected defect.
On the other hand, for the statistical approach, the best result is 80.20\%, which is by generated by \textit{HPIV}.
Similar to the edge detection feature extractor, the statistical approach attains an average classification result of $\textgreater$75\%.
The number of features required to represent each image for the handcrafted features is listed in Table~\ref{table:featureSize}.
It is observed that, although some of the feature sizes are very small (i.e., LBP[32 32] that has 59 features/ image), the classification accuracy is considered satisfactory (i.e., achieves an average accuracy of 77\%).

%%%%%%%%%%%%%%%%%%%%%%%%%%%%%%%%%%%%%%%%%%%%%%%%%%%%%%
\setlength{\tabcolsep}{5pt}

 \begin{table*}[t!]
 \begin{center}
 \caption{Classification accuracy (\%) by adopting various of edge detection features and evaluated on different types of classifiers}
 \label{table:resultEdge}
 \begin{tabular}{cccccccc}
\hline
\noalign{\smallskip}
 
\multicolumn{2}{c}{Classifier} 
 & Canny
 & Prewitt
 & Sobel
 & Roberts
 & LoG
 & ApproxCanny\\
 \hline
\noalign{\smallskip}

\multirow{3}{*}{Decision Tree}
& Fine Tree
    & 68.90
    & 76.40
    & 76.80
    & 72.00
    & 71.00
    & 80.00\\
\noalign{\smallskip}

& Medium Tree
    & 76.90
    & 79.30
    & 78.30
    & 77.60
    & 77.60
    & 80.00\\
\noalign{\smallskip}

& Coarse Tree
    & 78.90
    & 79.80
    & 79.10
    & 79.00
    & 79.30
    & 80.00\\
 \hline
\noalign{\smallskip}

\multirow{6}{*}{SVM}
& Linear SVM
    & 80.00
    & 79.90
    & 79.80
    & 78.90
    & 80.00
    & 80.20 \\
\noalign{\smallskip}

& Quadratic SVM
    & 78.90
    & 77.90
    & 77.20
    & 76.30
    & 79.00
    & 80.10 \\
\noalign{\smallskip}

& Cubic SVM
    & 75.80
    & 76.70
    & 75.80
    & 76.10
    & 76.30
    & 79.80 \\
\noalign{\smallskip}

& Fine Gaussian SVM
    & 80.00
    & 80.00
    & 80.00
    & 80.00
    & 80.00
    & \textbf{83.50} \\
\noalign{\smallskip}

& Medium Gaussian SVM
    & 80.00
    & 80.00
    & 80.00
    & 80.00
    & 80.00
    & 83.40\\
\noalign{\smallskip}

& Coarse Gaussian SVM
    & 80.00
    & 80.00
    & 80.00
    & 80.00
    & 80.00
    & 81.30\\
\noalign{\smallskip}

 \hline

\multirow{6}{*}{KNN}
& Fine KNN
    & 60.40
    & 67.90
    & 76.90
    & 79.00
    & 80.00
    & 80.10 \\
\noalign{\smallskip}

& Medium KNN
    & 79.60
    & 80.00
    & 80.00
    & 80.00
    & 80.00
    & 80.00 \\
\noalign{\smallskip}

& Coarse KNN
    & 80.00
    & 80.00
    & 80.00
    & 80.00
    & 80.00
    & 80.00 \\
\noalign{\smallskip}

& Cosine KNN
    & 79.50
    & 79.60
    & 79.40
    & 79.70
    & 79.80
    & 83.20 \\
\noalign{\smallskip}

& Cubic KNN
    & 79.60
    & 80.00
    & 80.00
    & 80.00
    & 80.00
    & 80.00\\
\noalign{\smallskip}

& Weighted KNN
    & 77.90
    & 79.90
    & 80.00
    & 80.00
    & 80.00
    & 80.00\\
\noalign{\smallskip}

 \hline

\multirow{5}{*}{Ensemble}
& Boosted Trees
    & 79.90
    & 80.00
    & 79.70
    & 79.90
    & 79.80
    & 79.90 \\
\noalign{\smallskip}

& Bagged Trees
    & 80.00
    & 80.00
    & 80.00
    & 77.50
    & 80.00
    & 80.30 \\
\noalign{\smallskip}

& Subspace Discriminant
    & 73.90
    & 73.80
    & 72.70
    & 73.70
    & 74.40
    & 81.20 \\
\noalign{\smallskip}

& Subspace KNN
    & 73.80
    & 79.90
    & 79.90
    & \textbf{80.10}
    & 80.00
    & 79.90 \\
\noalign{\smallskip}

& RUSBoosted Trees
    & 50.70
    & 51.40
    & 52.10
    & 66.30
    & 51.10
    &79.90\\
 \hline
  \hline
 \noalign{\smallskip}

\multicolumn{2}{c}{Average} 
    & 75.74
    & 76.25
    & 76.42
    & 77.81
    & 77.57
    & \textbf{80.64} \\
 \hline
\noalign{\smallskip}

 \end{tabular}
 \end{center}
 \end{table*}

\setlength{\tabcolsep}{5pt}

 \begin{table*}[t!]
 \begin{center}
 \caption{Classification accuracy (\%) by adopting various of statistical approaches and evaluated on different types of classifiers}
 \label{table:resultHist}
 \begin{tabular}{cccccccc}
\hline
\noalign{\smallskip}
 
\multicolumn{2}{c}{Classifier} 
 & HPIV
 & HOG[8 8]			
 & HOG[10 10]	
 & LBP[8 8]
 & LBP[16 16]
 & LBP[32 32]\\
 \hline
\noalign{\smallskip}

\multirow{3}{*}{Decision Tree}
& Fine Tree
    & 74.80	
    & 70.20
    & 70.00
    & 68.80
    & 69.80	
    & 71.70
\\
\noalign{\smallskip}

& Medium Tree
    & 79.00
    & 76.80
    & 77.30
    & 76.60	
    & 76.50
    & 76.90
\\
\noalign{\smallskip}

& Coarse Tree
    & 80.00	
    & 77.10
    & 78.70
    & 79.50
    & 78.90	
    & 78.70
\\
 \hline
\noalign{\smallskip}

\multirow{6}{*}{SVM}
& Linear SVM
    & 79.50
    & 80.00
    & 80.00
    & \textbf{80.10}
    & 80.00
    & 80.00
 \\
\noalign{\smallskip}

& Quadratic SVM
    & 77.60
    & 79.50
    & 79.50
    & 78.10
    & 79.80
    & 78.70
 \\
\noalign{\smallskip}

& Cubic SVM
    & 71.00
    & 75.80
    & 76.30
    & 76.20
    & 77.20
    & 74.20
 \\
\noalign{\smallskip}

& Fine Gaussian SVM
    & \textbf{80.20}
    & 80.00
    & 80.00
    & 80.00
    & 80.00
    & 80.00 \\
\noalign{\smallskip}

& Medium Gaussian SVM
    & 80.10
    & 80.00
    & 80.00
    & 80.00
    & 80.00
    & \textbf{80.10}\\
\noalign{\smallskip}

& Coarse Gaussian SVM
    & 80.00
    & 80.00
    & 80.00
    & 80.00
    & 80.00
    & 80.00\\
\noalign{\smallskip}

 \hline

\multirow{6}{*}{KNN}
& Fine KNN
    & 68.90
    & 69.90
    & 69.60
    & 74.00
    & 72.80
    & 69.50
\\
\noalign{\smallskip}

& Medium KNN
    & 79.50	
    & 79.70
    & 79.70
    & 80.00	
    & 79.90
    & 79.00
\\
\noalign{\smallskip}

& Coarse KNN
    & 80.00
    & 80.00
    & 80.00
    & 80.00
    & 80.00
    & 80.00 \\
\noalign{\smallskip}

& Cosine KNN
    & 80.00
    & 79.30
    & 79.80
    & 79.90
    & 79.80
    & 79.80
\\
\noalign{\smallskip}

& Cubic KNN
    & 79.50
    & 79.60
    & 79.80	
    & 79.90	
    & \textbf{80.10}
    & 79.20
\\
\noalign{\smallskip}

& Weighted KNN
    & 79.00	
    & 79.20
    & 78.90
    & 79.90	
    & 79.50
    & 78.60
\\
\noalign{\smallskip}

 \hline

\multirow{5}{*}{Ensemble}
& Boosted Trees
    & 80.00
    & 79.90
    & 79.90	
    & 79.80	
    & 79.90	
    & 80.00
 \\
\noalign{\smallskip}

& Bagged Trees
    & 80.10
    & 79.40
    & 79.70
    & 79.80
    & 79.90
    & 79.80
 \\
\noalign{\smallskip}

& Subspace Discriminant
    & 79.50	
    & 77.40
    & 79.40	
    & 73.90
    & 79.60
    & 80.00
\\
\noalign{\smallskip}

& Subspace KNN
    & 75.40	
    & 76.30
    & 77.00
    & 79.20
    & 79.10	
    & 79.30
\\
\noalign{\smallskip}

& RUSBoosted Trees
    & 58.20
    & 55.40
    & 53.50
    & 53.80
    & 55.90
    & 55.00
\\
 \hline
 \hline
\noalign{\smallskip}

\multicolumn{2}{c}{Average}
    & 77.12
    & 76.42
    & 76.96
    & 76.06
    & \textbf{77.44}
    & 77.03
\\
 \hline
\noalign{\smallskip}

 \end{tabular}
 \end{center}
 \end{table*}

%%%%%%%%%%%%%%%%%%%%%%%%%%%%%%%%%%%%%%%%%%%%%%%%%%%%%%
\setlength{\tabcolsep}{5pt}

 \begin{table}[t!]
 \begin{center}
 \caption{Feature length per image for handcrafted features (i.e., edge detector and statistical approach)}
 \label{table:featureSize}
 \begin{tabular}{lcc}
\cline{2-3}
\noalign{\smallskip}

& \multirow{2}{*}{Feature Extractor}
& Feature size
\\

& 
& per image
\\ 
\hline
\noalign{\smallskip}
\multirow{6}{*}{Edge Detector}
& Canny
& 1600
\\ 
\noalign{\smallskip}
& Prewitt
& 1600
\\ 
\noalign{\smallskip}
& Sobel
& 1600
\\ 
\noalign{\smallskip}
& Roberts
& 1600
\\ 
\noalign{\smallskip}
& LoG
& 1600
\\ 
\noalign{\smallskip}
& Approxcanny
& 1600
\\ 
\noalign{\smallskip}
\hline
\multirow{6}{*}{Statistical Approach}
& HPIV
& 256
\\ 
\noalign{\smallskip}

& HOG [8 8]
& 576
\\ 
\noalign{\smallskip}
& HOG [10 10]
& 324
\\ 
\noalign{\smallskip}
& LBP [8 8]
& 1475
\\ 
\noalign{\smallskip}
& LBP [16 16]
& 236
\\ 
\noalign{\smallskip}
& LBP [32 32]
& 59
\\
\hline
\noalign{\smallskip}
 \end{tabular}
 \end{center}
 \end{table}

%%%%%%%%%%%%%%%%%%%%%%%%%%%%%%%%%%%%%%%%%%%%%%%%%%%%%%
%%%%%%%%%%%%%%%%%%%%%%%%%%%%%%%%%%%%%%%%%%%%%%%%%%%%%%

\begin{figure} [t!]
    \centering
    \includegraphics[width = 0.8\linewidth]{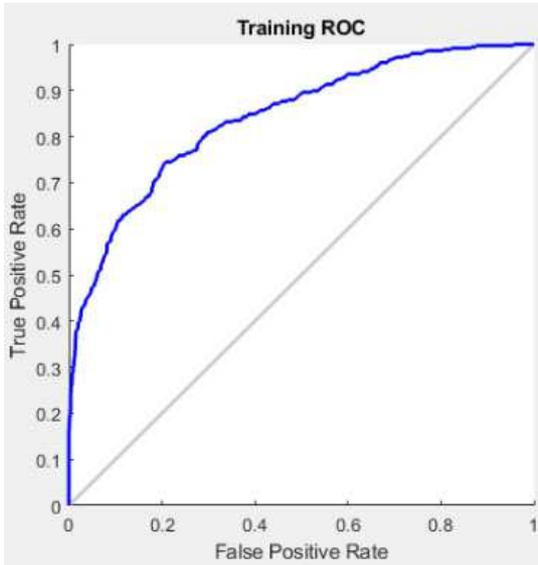}
    \caption{ROC for the training data}
    \label{fig:trainROC}
\end{figure}

\begin{figure} [t!]
    \centering
    \includegraphics[width = 0.8\linewidth]{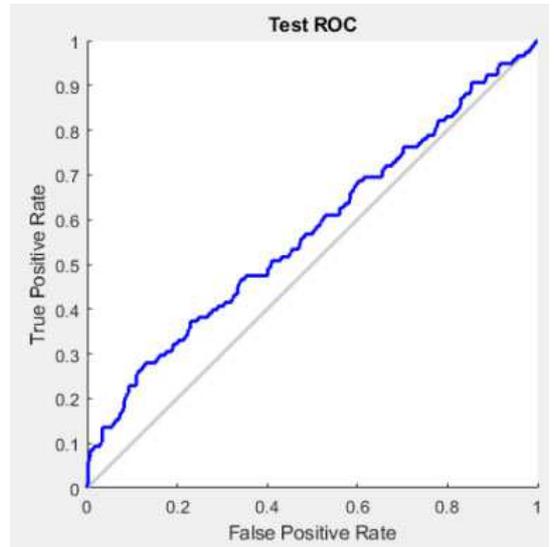}
    \caption{ROC for the testing data}
    \label{fig:testROC}
\end{figure}

On another note, the classification performance of a method that adopts \textit{ApproxCanny} as the pre-processing, then further processing by ANN is reported in Table~\ref{table:resultANN}.
The highest classification accuracy exhibited is 82.49\%, with the train/ test split of 75/25 and the number of neurons in the hidden layer set to 50.
Figure~\ref{fig:trainROC} and Figure~\ref{fig:testROC} illustrate the Receiver Operating Characteristic (ROC) to verify the quality of the binary classifiers for the ANN training and testing progress. 
The ROC curve plots the true positive rate (TPR) against the false positive rate (FPR) at multiple threshold instances. 
During the training process, most of the points are located at the upper left region, which indicates that a good classifier has been modeled.

%%%%%%%%%%%%%%%%%%%%%%%%%%%%%%%%%%%%%%%%%%%%%%%%%%%%%%
%%%%%%%%%%%%%%%%%%%%%%%%%%%%%%%%%%%%%%%%%%%%%%%%%%%%%%
\setlength{\tabcolsep}{5pt}

 \begin{table}[t!]
 \begin{center}
 \caption{Classification accuracy (\%) by adopting top features extractor (\textit{ApproxCanny}) and further process using ANN with different number of neuron in the hidden layer on various train-test partition}
 \label{table:resultANN}
 \begin{tabular}{cccccccc}
\hline

\multirow{2}{*}{No. of}
&\multicolumn{6}{c}{Train/ test partition}\\
\cline{2-7}

\noalign{\smallskip}
neuron
    & 70/ 30
    & 75/ 25
    & 80/ 20
    & 85/ 15
    & 90/ 10
    & 95/ 5
\\ 

\hline
\noalign{\smallskip}
 60
    & 78.82
    & 78.96
    & 78.53
    & 73.88
    & 79.41
    & 77.31
\\ 

\hline
\noalign{\smallskip}
 50
    & 81.91
    & \textbf{82.49}
    & 78.74
    & 79.49
    & 78.57
    & 78.15
\\ 

\hline
\noalign{\smallskip}
 40
& 81.07
& 79.97
& 81.26
& 81.18
& 78.57
& 79.83
\\

\hline
\noalign{\smallskip}
30
& 80.65
& 80.30
& 80.42
& 80.34
& 78.15
& 76.47
\\ 
\hline
\noalign{\smallskip}
 \end{tabular}
 \end{center}
 \end{table}

%%%%%%%%%%%%%%%%%%%%%%%%%%%%%%%%%%%%%%%%%%%%%%%%%%%%%%
%%%%%%%%%%%%%%%%%%%%%%%%%%%%%%%%%%%%%%%%%%%%%%%%%%%%%%

Table~\ref{table:cf1} and Table~\ref{table:cf2} tabulate the confusion matrix that describes the performance of the test data, which correspond to the highest results yielded by the \textit{ApproxCanny} handcrafted features (i.e., accuracy of 83.5\%) and ANN (i.e., accuracy of 82.49\%).
For the scenario with \textit{ApproxCanny} edge detector as the feature extractor, all the images (i.e., 2376 samples) are tested as it is implementing a 5-fold cross validation strategy.
As for the ANN, the train/ test split is 75/ 25.
Thus, there will be a total of 594 testing data.
Since $\sim$80\% of the images in the dataset do not contain any defects (i.e., 1903 images do not contain any defect and 475 images contain at least one defect), the testing accuracy for images with no defects are significantly higher than with defect images.

%%%%%%%%%%%%%%%%%%%%%%%%%%%%%%%%%%%%%%%%%%%%%%%%%%%%%%
%%%%%%%%%%%%%%%%%%%%%%%%%%%%%%%%%%%%%%%%%%%%%%%%%%%%%%
\setlength{\tabcolsep}{5pt}

 \begin{table}[tb]
 \begin{center}
 \caption{Confusion matrix of the \textit{ApproxCanny} edge detector as feature extractor, which evaluate using 5-fold cross validation}
 \label{table:cf1}
 \begin{tabular}{lcccccc}
    
    \noalign{\smallskip}
    \cline{3-4} 
    \noalign{\smallskip}
    
    & & \multicolumn{2}{c}{Predicted}\\
    
    \noalign{\smallskip}
    \cline{3-4} 
    \noalign{\smallskip}
    
    & & No defect	 & Has defect	\\
    
    \noalign{\smallskip}
    \hline
    \noalign{\smallskip}
    
    \multirow{2}{*}{Actual}
    
    & No defect  
    & 1836	 & 65		\\
    
    & Has defect
    & 328	 & 147 \\
    
    \hline
 \end{tabular}
 \end{center}
 \end{table}

%%%%%%%%%%%%%%%%%%%%%%%%%%%%%%%%%%%%%%%%%%%%%%%%%%%%%%
%%%%%%%%%%%%%%%%%%%%%%%%%%%%%%%%%%%%%%%%%%%%%%%%%%%%%%
\setlength{\tabcolsep}{5pt}

 \begin{table}[tb]
 \begin{center}
 \caption{Confusion matrix of the \textit{ApproxCanny} edge detector and ANN as feature extractor, which evaluate on 25\% images from the dataset}
 \label{table:cf2}
 \begin{tabular}{lcccccc}
    
    \noalign{\smallskip}
    \cline{3-4} 
    \noalign{\smallskip}
    
    & & \multicolumn{2}{c}{Predicted}\\
    
    \noalign{\smallskip}
    \cline{3-4} 
    \noalign{\smallskip}
    
    & & No defect	 & Has defect	\\
    
    \noalign{\smallskip}
    \hline
    \noalign{\smallskip}
    
    \multirow{2}{*}{Actual}
    
    & No defect  
    & 478	 & 97		\\
    
    & Has defect
    & 7	 & 12 \\
    
    \hline
 \end{tabular}
 \end{center}
 \end{table}
\section{Conclusion}
\label{sec:conclusion}

This study proposed a simple yet efficient solution to automatically classify the tick bite defects on calf leather.
Comprehensive experiments and analyses have been carried out to demonstrate the effectiveness of the proposed algorithm.
Overall, promising results are obtained by adopting both the handcrafted and data-driven features.
In particular, the handcrafted features include the edge detectors (i.e., edge detectors and histograms) and the data-driven approach (i.e., artificial neural network). 
Multiple supervised classifiers are exploited to determine if the input image contain defects or not.
For future works, other types of defects such as open cuts, closed cuts, wrinkles and scabies can be examined with similar procedures.
In addition, instead of collecting the sample images from a single type of leather, the hides of other animals like crocodile, sheep, monitor lizard, etc. can also be considered.
Furthermore, convolutional neural network (CNN) can be utilized to learn features as well as to classify input data.
Popular CNN architecture such as AlexNet, GoogLeNet, ResNet-50, VGG-16 can be modified to discover the meaningful local features in the leather images.

\section*{Acknowledgments}
This work was funded by Ministry of Science and Technology (MOST) (Grant Number: MOST 107-2218-E-035-016-), National Natural Science Foundation of China (No. 61772023) and Natural Science Foundation of Fujian Province (No. 2016J01320) .
The authors would also like to thank Hsiu-Chi Chang, Chien-An Wu, Cheng-Yan Yang, and Wen-Hung Lin for their assistance in the sample preparation, experimental setup  and image acquisition.
%The authors would also like to thank ..... for their assistance in the sample preparation, experimental setup and image acquisition.

\bibliography{mybibfile}

\end{document}